\documentclass{article}

\usepackage{arxiv}

\usepackage[utf8]{inputenc} 
\usepackage[T1]{fontenc}    
\usepackage{hyperref}       
\usepackage{url}            
\usepackage{booktabs}       
\usepackage{amsfonts}       
\usepackage{nicefrac}       
\usepackage{microtype}      
\usepackage{lipsum}
\usepackage{graphicx}
\usepackage{amsmath}
\graphicspath{ {./images/} }

\title{From Quantifying Vagueness To Pan-nifty-ism}

\author{
 Natesh Ganesh \\
  ITL, App. \& Comp. Mathematics Division, NIST \\
  Dept. of Physics, University of Colorado, Boulder, CO\\
   \texttt{natesh.ganesh@colorado.edu} \\
  \\
}

\newcommand{\RomanNumeralCaps}[1]
    {\MakeUppercase{\romannumeral #1}}

\begin{document}
\maketitle

\begin{abstract}
In this short paper, we will introduce a simple model for quantifying  philosophical \emph{vagueness}. We will then discuss some of the implications of this model including the conditions under which the quantification of $\mathcal{X}$=`nifty' leads to pan-nifty-ism. Understanding this leads to an interesting insight - the reason a framework to quantify consciousness like Integrated Information Theory (IIT) implies (forms of) panpsychism is because there is favorable structure already implicitly encoded in the construction of the quantification metric.
\end{abstract}


\section{A Model for Quantification in  Vagueness}
\textit{Vagueness} in philosophy is standardly defined by the possession of borderline cases \cite{VagueSEP} - \textit{``For example, `tall' is vague because a man who is 1.8 meters in height is neither clearly tall nor clearly non-tall. No amount of conceptual analysis or empirical investigation can settle whether a 1.8 meter man is tall. Borderline cases are inquiry resistant.''} The opposite of this would be the idea of sharpness characterized by having no borderline cases. In this paper, we will construct a simple yet powerful mathematical model of how borderline cases might arise in our epistemic pursuits, and use it better understand vagueness and attempts to quantify it. This is important since many important concepts like consciousness, agency, sentience could all be vague and there is growing interest in their quantification. There has been interesting work in the area of quantifying vagueness and would allow us access further mathematical tools to address this area of research \cite{Zimmerman}, \cite{Lassiter}. Furthermore it will allows us to explore whether things that might appear to be vague initially might get sharpened through an iterative process of observations and model-building.

Let there be a property $\mathcal{P}$ (the use of the word property is more colloquial and not to be confused with any rigorous definition from philosophy. From the example above, 'tall' is a property $\mathcal{P}$, and we wish to determine if a person/system of height 1.8 meters has/exhibits the property $\mathcal{P}:$ `tall') that we wish to study and quantify. For this $\mathcal{P}$, our model to will be characterized by the tuple  $\{ \mathcal{A}, (\{\mathcal{S}\}_C, \{\mathcal{S}\}_{\bar{C}}, \{\mathcal{S}\}_B), \{ \hat{\mathcal{O}}_i\}, \phi \}_{\mathcal{P}}$ where we have

\begin{itemize}
    \item An observer $\mathcal{A}$ with finite observational capabilities.
    \item A collection of systems that are apriori \textit{identified} by $\mathcal{A}$ as a clear case of exhibiting $\mathcal{P}$, a clear case of not exhibiting $\mathcal{P}$ and borderline cases that are organized into the sets $\{\mathcal{S}\}_C$, $\{\mathcal{S}\}_{\bar{C}}$ and $\{\mathcal{S}\}_B$ respectively.
    \item A set of measurement/observation mechanisms $\{ \hat{\mathcal{O}}_i\}$ utilized by $\mathcal{A}$ to generate a D-dimensional vector $\bar{x}^{\mathcal{S}}: \mathcal{S} \xrightarrow{\{ \hat{\mathcal{O}}_i\}} \bar{x}^{\mathcal{S}} \in \mathcal{R}^D$ of observables. The observer looks to determine if system $\mathcal{S}$ exhibits a property $\mathcal{P}$ using $\bar{x}^{\mathcal{S}}$.
    \item A function $\phi : \mathcal{R}^D \rightarrow \mathcal{R}^N$ to \textit{faithfully} (defined later in this section) quantify $\mathcal{P}$, that will map the $D$-dimensional $\bar{x}^{\mathcal{S}}$ to a $N$-dimensional vector $\phi(\bar{x}^{\mathcal{S}})$. The existence of a trivial faithful $\phi$ is discussed later.
    \item A system of interest $\mathcal{S}$ which for which we intend to \textit{determine} whether or not it exhibits property $\mathcal{P}$ using $\phi$.
\end{itemize}

The property $\mathcal{P}$ that is of interest of study is implicitly described by identifying the systems that make up $\{\mathcal{S}\}_C$, $\{\mathcal{S}\}_{\bar{C}}$ and $\{\mathcal{S}\}_B$. We are not sure whether it is possible to identify some systems apriori as borderline cases of $\mathcal{P}$ to form the elements of $\{\mathcal{S}\}_B$, as opposed to simply determine them to be borderline cases using $\phi$. It is possible to view the elements of $\{\mathcal{S}\}_B$ as test cases for $\phi$ constructed using $\{\mathcal{S}\}_C$ and $\{\mathcal{S}\}_{\bar{C}}$ (this will be further evident below when discussing a faithful $\phi$). The strength of the model presented here lies in the generality of it's application and results presented in this work do not depend on these unresolved questions with respect to $\{\mathcal{S}\}_B$. If $\{\mathcal{S}\}_B$ could be an empty or non-empty set, then we can use it to form a (weaker) definition of sharp and vague based on apriori observer identification. For a \textit{sharp} $\mathcal{P}$ with no apriori borderline cases $\{\mathcal{S}\}_B =\{ \}$ and $\{\mathcal{S}\}_B \neq \{ \}$ for a vague $\mathcal{P}$. The function $\phi$ can then be constructed on the basis of the current elements in these three sets (though the necessity to identify and utilize $\{\mathcal{S}\}_{\bar{C}}$ in this construction is contested and will be the focus of our discussion later in the paper). Without loss of generality, we will deal with the case of a scalar $\phi(\bar{x}^{\mathcal{S}})$. This can be generalized for $N$-dimensional vector later if needed using appropriate distance metrics. In this framework, will call $\phi$ (with $\alpha_{\phi} \leq \phi \leq \beta_{\phi})$ a \textit{faithful} quantification of $\mathcal{P}$ if 
\begin{itemize}
    \item There is at least one system $\mathcal{S} \in \{\mathcal{S}\}_C$ for which we define $\eta_0= \phi(\bar{x}^{\mathcal{S}})$ such that $\phi(\bar{x}^{\mathcal{S}}) \geq \eta_0$ for all current elements in $\{\mathcal{S}\}_C$ i.e. $ \phi(\bar{x}^{\mathcal{S}})$ is in the interval $[\eta_0,\beta_{\phi}]$ for all $\mathcal{S} \in \{\mathcal{S}\}_C$.
    \begin{center}
    $\eta_0 = \min\limits_{\{ \forall \mathcal{S}: \mathcal{S} \in \{\mathcal{S}\}_C, \mathcal{S} \xrightarrow{\{ \mathcal{O}_i\}} \bar{x}^{\mathcal{S}}  \}}  \phi(\bar{x}^{\mathcal{S}})$
    \end{center}
    
    \item There is at least one system $\mathcal{S} \in \{\mathcal{S}\}_{\bar{C}}$ for which we define $\gamma_0= \phi(\bar{x}^{\mathcal{S}})$ such that $\phi(\bar{x}^{\mathcal{S}}) \leq \gamma_0$ for all current elements in $\{\mathcal{S}\}_{\bar{C}}$ i.e. $ \phi(\bar{x}^{\mathcal{S}})$ is in the interval $[\alpha_{\phi},\gamma_0]$ for all $\mathcal{S} \in \{\mathcal{S}\}_{\bar{C}}$.
    \begin{center}
    $\gamma_0 = \max\limits_{\{ \forall \mathcal{S}: \mathcal{S} \in \{\mathcal{S}\}_{\bar{C}}, \mathcal{S} \xrightarrow{\{ \mathcal{O}_i\}} \bar{x}^{\mathcal{S}}  \}}  \phi(\bar{x}^{\mathcal{S}})$
    \end{center}
    
    \item As a consequence of the above two conditions, we would have that $\gamma_0 < \phi(\bar{x}^{\mathcal{S}}) < \eta_0$ for all the apriori borderline cases $\mathcal{S} \in \{\mathcal{S}\}_B$.
\end{itemize}

The above conditions also allows us to produce a stronger definition of sharp and vague. $\mathcal{P}$ is sharp using $\phi$ if $\gamma_0=\eta_0$ and there is no interval for the borderline cases, and $\mathcal{P}$ is vague if $\gamma_0 \neq \eta_0$ and there is a interval for borderline cases. We can also show that for a given $\{\mathcal{S}\}_C$, $\{\mathcal{S}\}_{\bar{C}}$ and $\{\mathcal{S}\}$ for $\mathcal{P}$, a trivial yet faithful $\phi$ that satisfies the conditions above always exists, and can be constructed as 

\begin{equation}
\phi(x^{\mathcal{S}_i}) =
\left\{
	\begin{array}{ll}
		\eta_0=\beta_{\phi}  & \mbox{if }\mathcal{S}_i \in \{\mathcal{S}\}_{C} \\
	    \gamma_0=\alpha_{\phi}  & \mbox{if }\mathcal{S}_i \in \{\mathcal{S}\}_{\bar{C}} \\
	\end{array}
\right. \nonumber 
\end{equation}

We should also note here that the values of $\eta_0$ and $\gamma_0$ are not fixed and only represent a snapshot of the current knowledge that observer $\mathcal{A}$ has with respect to apriori cases given by $\{\mathcal{S}\}_C$, $\{\mathcal{S}\}_{\bar{C}}$ and $\{\mathcal{S}\}_B$ and the construction of $\phi$ based on that. If a new system that is apriori identified as a clear case of either exhibiting or not exhibiting $\mathcal{P}$ is presented to $\mathcal{A}$ (we exclude $\{\mathcal{S}\}_B $ for now, since it an open question as to whether we can apriori identify borderline cases), the observer might have to make updates to both $\phi$, the values $\eta_0$ and $\gamma_0$ and the corresponding intervals to account for that. This identification of new elements of $\{\mathcal{S}\}_C$ and $\{\mathcal{S}\}_{\bar{C}}$ represent a positive since it allows for an improvement in the implicit description of $\mathcal{P}$ by $\mathcal{A}$. This improvement in the identification of $\mathcal{P}$ through new example of clear cases could result in the reduction of the interval $(\gamma_0,\eta_0)$. As a result of this shrinkage, systems that were previously determined to be borderline might now belong to either $\{\mathcal{S}\}_C$ or $\{\mathcal{S}\}_{\bar{C}}$ - achieving a \textit{sharpening} of $\mathcal{P}$. In an extreme case, constant updates to $\{\mathcal{S}\}_C$ and $\{\mathcal{S}\}_{\bar{C}}$ results in making $\gamma_0=\eta_0$ and reduces $\{\mathcal{S}\}_B= \{ \}$. On the other hand, if we were to obtain new examples that are identified as borderline cases of $\mathcal{P}$, these new elements of $\{\mathcal{S}\}_B$ might need an expansion of the interval $(\gamma_0,\eta_0)$ and absorb elements that were previously viewed as clear cases. We would once again need to make updates to $\phi$ to reflect these changes. This latter phenomenon can be viewed as a \textit{vaguening} of $\mathcal{P}$. Both sharpening and vaguening represent the effect of the updates to $\{\mathcal{S}\}_C$, $\{\mathcal{S}\}_{\bar{C}}$ and $\{\mathcal{S}\}_B$. We will next present the 1st main proposition of this paper built using the framework from above.

\textbf{Proposition-$\RomanNumeralCaps{1}$}: For the property $\mathcal{P}$ in the framework characterized by $\{ \mathcal{A}, (\{\mathcal{S}\}_C, \{\mathcal{S}\}_{\bar{C}}, \{\mathcal{S}\}_B), \{ \hat{\mathcal{O}}_i\}, \phi \}_{\mathcal{P}}$, we propose that

\begin{itemize}
\item[(1.1)] A system $\mathcal{S}$ exhibits $\mathcal{P}$ if and only if $\phi(\bar{x}^{\mathcal{S}}) \geq \eta_0$. We know that $\eta_0$ exists for a faithful $\phi$ by definition.

\item[(1.2)] A system $\mathcal{S}$ does not exhibit $\mathcal{P}$ if and only if 
$\phi(\bar{x}^{\mathcal{S}}) \leq \gamma_0$. We know that $\gamma_0$ exists for a faithful $\phi$ by definition.

\item[(1.3)] A consequence of (1.1) and (1.2) above would be that a system $\mathcal{S}$ is considered to be a borderline case of $\mathcal{P}$ if $\phi(\bar{x}^{\mathcal{S}})$ is in the interval $(\gamma_0,\eta_0)$.
\end{itemize}

It is also important to note that once we use $\phi$ to determine whether the new system $\mathcal{S}$ either clearly exhibits or does not exhibit $\mathcal{P}$ or is a borderline case, the system should not be used to update the sets $\{\mathcal{S}\}_C$, $\{\mathcal{S}\}_{\bar{C}}$ and $\{\mathcal{S}\}_B$ respectively. Only systems for which we make some type of apriori identification with respect to $\mathcal{P}$ (independent of $\phi$) can be used to update the corresponding sets and $\phi$ (if needed). The addition of these new systems $\mathcal{S}$ to $\{\mathcal{S}\}_C$, $\{\mathcal{S}\}_{\bar{C}}$ and $\{\mathcal{S}\}_B$ would result in a false sharpening or vaguening of $\mathcal{P}$.

In addition to the requirement that $\phi$ be a faithful quantification, it would also be favorable for $\phi$ to quantitatively distinguish between different elements in a manner that would match with our expectation for $\mathcal{P}$. For example, consider $\mathcal{S}_1$ and $\mathcal{S}_2$ in $\{\mathcal{S}\}_C$ with $\phi$ and $\eta_0=2$. With $\phi(\bar{x}^{\mathcal{S}_1}) = 4$ and $\phi(\bar{x}^{\mathcal{S}_2}) = 8$, both $\mathcal{S}_1$ and $\mathcal{S}_2$ satisfy the condition $\phi(\bar{x}^{\mathcal{S}_i}) \geq 2$ but for two very different values of $\phi$, with $\phi(\bar{x}^{\mathcal{S}_2}) = 2 \times \phi(\bar{x}^{\mathcal{S}_2})$. This is especially favorable if we are identifying $\mathcal{S}_2$ as exhibiting $\mathcal{P}$ \textit{more} than $\mathcal{S}_1$. The difference in the values of $\phi(\bar{x}^{\mathcal{S}_i})$ thus allows us to construct a distance metric based on $(\phi(\bar{x}^{\mathcal{S}_i})-\eta_0)$ and quantify how much more or less a particular system $\mathcal{S}_i$ exhibits property $\mathcal{P}$. This becomes very evident from the construction the following function $\phi^{\prime}$ - 

\begin{equation}
\phi^{\prime}(x^{\mathcal{S}_i}) =
\left\{
	\begin{array}{ll}
		1  & \mbox{if } \phi(x^{\mathcal{S}_i}) \geq \eta_0 \\
		0 & \mbox{if } \phi(x^{\mathcal{S}_i}) < \eta_0
	\end{array}
\right. \nonumber 
\end{equation}

\noindent We can see that $\phi^{\prime}(x^{\mathcal{S}_i})$ can also be used faithfully to describe the same property $\mathcal{P}$ as $\phi^{\prime}(x^{\mathcal{S}}) \geq \eta^{\prime}_0=1$ for all $\mathcal{S}_i \in \{\mathcal{S}\}_C$. However since  all the elements in $\{\mathcal{S}\}_C$ now have the same value of $\phi^{\prime}(x^{\mathcal{S}_i}) = 1$ which is not useful if one would like to distinguish between $\mathcal{S}_1$ and $\mathcal{S}_2$. While we have already that a faithful $\phi$ always exists, the existence of favorable $phi$ to appropriately distinguish between systems is an open question.

In this section, we have been discussing vagueness for an observer $\mathcal{A}$ as having $\{\mathcal{S}\}_B \neq \{\}$, as well as the expansion and shrinkage of vagueness. We will now explore vagueness arising in conditions with multiple observers (we will use two without loss of generality). For property $\mathcal{P}$, the two observers $\mathcal{A}_1$ and $\mathcal{A}_2$ are characterized by the two tuples - $\{ \mathcal{A}_1, (\{\mathcal{S}\}^{\mathcal{A}_1}_C, \{\mathcal{S}\}^{\mathcal{A}_1}_{\bar{C}}, \{\mathcal{S}\}^{\mathcal{A}_1}_B), \{ \hat{\mathcal{O}}_i\}^{\mathcal{A}_1}, \phi_1 \}_{\mathcal{P}}$ and $\{ \mathcal{A}_2, (\{\mathcal{S}\}^{\mathcal{A}_2}_C, \{\mathcal{S}\}^{\mathcal{A}_2}_{\bar{C}}, \{\mathcal{S}\}^{\mathcal{A}_2}_B), \{ \hat{\mathcal{O}}_i\}^{\mathcal{A}_1}, \phi_2 \}_{\mathcal{P}}$ respectively. While different observers could have different observational capabilities that affect how they quantify $\mathcal{P}$, we will assume $\{ \hat{\mathcal{O}}_i\}^{\mathcal{A}_1}= \{ \hat{\mathcal{O}}_i\}^{\mathcal{A}_2}$ to instead focus on differences elsewhere and their corresponding effects. Let $\mathcal{A}_1$ and $\mathcal{A}_2$ disagree as to the elements in $\{\mathcal{S}\}_C$, $\{\mathcal{S}\}_{\bar{C}}$ and/or $\{\mathcal{S}\}_B$. The difference in these sets could result in differences in $\phi$, $\eta_0$ and/or $\gamma_0$ values. As a result of this, for the same system $\mathcal{S}$ we could have $\phi_1(\bar{x}^{\mathcal{S}}) \geq \eta_0^{\mathcal{A}_1}$ and $\phi_2(\bar{x}^{\mathcal{S}}) \leq \gamma_0^{\mathcal{A}_2}$. Thus observers $\mathcal{A}_1$ and $\mathcal{A}_2$ take opposite positions on $\mathcal{S}$ with respect to $\mathcal{P}$. This is a form of vagueness in $\mathcal{P}$ with respect to $\mathcal{S}$ arising due to differences between the observers, since the two observers by their identification of the clear cases are constructing different characterizations of $\mathcal{P}$ that might not always overlap. Interestingly even with $\{\mathcal{S}\}^{\mathcal{A}_1}_B)=\{ \}$ and $\{\mathcal{S}\}^{\mathcal{A}_2}_B)=\{ \}$ making $\mathcal{P}$ sharp when we consider the observers individually, there is a vagueness that emerges when we take them together. The path forward involves agreeing that both $\phi_1$ and $\phi_2$ are equally valid and the two observers are simple talking about different $\mathcal{P}$. If the observers agree that they are in fact talking about the same $\mathcal{P}$, the next step would be to view the two sets of $(\phi, \eta_0, \gamma_0$ as competing models and quantification metrics that are on equal footing with respect to their respective $(\{\mathcal{S}\}^{\mathcal{A}_1}_C, \{\mathcal{S}\}^{\mathcal{A}_1}_{\bar{C}}, \{\mathcal{S}\}^{\mathcal{A}_1}_B)$. Use of systems that the observers disagree upon to judge $\phi^{\mathcal{P}}_1$ and $\phi^{\mathcal{P}}_2$ would be fruitless and only reinforce the difference in what either observer describes as $\mathcal{P}$. If a system is identified apriori by both observers to belong to either $\{\mathcal{S}\}^{\mathcal{A}_1}_C$ or $\{\mathcal{S}\}^{\mathcal{A}_1}_{\bar{C}}$, then that system can be used to judge which of $\phi_1$ and $\phi_2$ is the better metric. The failed metric can either be abandoned or improved to account for this new system.

Now we explore a special case of the above in greater detail - assume that $\mathcal{A}_1$ and $\mathcal{A}_1$ agree on $\{\mathcal{S}\}_C$, but not on $\{\mathcal{S}\}_{\bar{C}}$. For $\mathcal{A}_1$, $ \{\mathcal{S}\}^{\mathcal{A}_1}_{\bar{C}} = \{ \}$ and $ \{\mathcal{S}\}^{\mathcal{A}_2}_{\bar{C}} \neq \{ \}$ for $\mathcal{A}_2$. Thus observer $\mathcal{A}_1$ is not identifying any system as a clear case of not satisfying $\mathcal{P}$, whereas the elements of $\{\mathcal{S}\}^{\mathcal{A}_2}_{\bar{C}}$ meet that criteria for $\mathcal{A}_2$. As before, they could construct different measures $\phi_1$ and $\phi_2$ but we will assume that $\phi_1=\phi_2=\phi_{\mathcal{P}}$. Since both observers share the same $\{\mathcal{S}\}_C$, we will have $\eta_0^{\mathcal{A}_1}=\eta_0^{\mathcal{A}_2}=\eta_0^{\mathcal{P}}$ and the condition to determine whether a new system $\mathcal{S}$ exhibits $\mathcal{P}$ - $\phi_{\mathcal{P}}(\bar{x}^{\mathcal{S}}) \geq \eta_0^{\mathcal{P}}$ is the same for both $\mathcal{A}_1$ and $\mathcal{A}_2$ i.e. $\mathcal{A}_1$ and $\mathcal{A}_2$ will always agree on which systems exhibit $\mathcal{P}$ using $\phi^{\mathcal{P}}$. This successful agreement between $\mathcal{A}_1$ and $\mathcal{A}_2$ might inspire additional confidence in the efficacy of $\phi_{\mathcal{P}}$ for $\mathcal{A}_1$ built under the assumption of $\{\mathcal{S}\}_{\bar{C}}=\{\}$. The differences between $\{\mathcal{S}\}^{\mathcal{A}_1}_{\bar{C}}$ and $\{\mathcal{S}\}^{\mathcal{A}_2}_{\bar{C}}$ however will result in different values of $\gamma^{\mathcal{A}_1}_0$ and $\gamma^{\mathcal{A}_2}_0$ and the corresponding interval in which a system is determined to not exhibit $\mathcal{P}$ based on $\phi_{\mathcal{P}}$. For $\phi_{\mathcal{P}}$, with $\alpha_{\phi} \leq \phi_{\mathcal{P}} \leq \beta_{\phi})$, we can define $\gamma^{\mathcal{A}_2}_0$ using $\{\mathcal{S}\}^{\mathcal{A}_2}_{\bar{C}} \neq \{ \}$. Thus for $\mathcal{A}_2$, a system $\mathcal{S}$ does not exhibit $\mathcal{P}$ if $\alpha_{\phi} \leq \phi_{\mathcal{P}} (\bar{x}^{\mathcal{S}}) \leq \gamma^{\mathcal{A}_2}_0 $. Since $\{\mathcal{S}\}^{\mathcal{A}_1}_{\bar{C}} = \{ \}$ for observer $\mathcal{A}_1$, we would have $\gamma^{\mathcal{A}_1}_0=\alpha_{\phi}$ by definition. This implies that a system $\mathcal{S}$ does not exhibit a $\mathcal{P}$ only when $\phi_{\mathcal{P}} (\bar{x}^{\mathcal{S}})= \alpha_{\phi}$, which is the lower bound of $\phi_{\mathcal{P}}$. As stated earlier in this section, we have the interval of $\phi_{\mathcal{P}}$ for which a system is determined to be a borderline case as $(\gamma_0,\eta_0)$. Since $\gamma^{\mathcal{A}_1}_0=\alpha_{\phi}$ for $\mathcal{A}_1$, the corresponding interval for borderline cases is $(\gamma^{\mathcal{A}_1}_0,\eta^{\mathcal{A}_1}_0)$. As a result, observer $\mathcal{A}_1$ will determine that any system $\mathcal{S}$ with $\phi_{\mathcal{P}}(\bar{x}^{\mathcal{S}_2})$ such that $\gamma^{\mathcal{A}_1}_0 < \phi_{\mathcal{P}}(\bar{x}^{\mathcal{S}_2}) < \eta^{\mathcal{A}_1}_0 $ as a borderline case of exhibiting $\mathcal{P}$. This would be in disagreement for those systems with $\phi_{\mathcal{P}} (\bar{x}^{\mathcal{S}}) \leq \gamma^{\mathcal{A}_2}_0$ that $\mathcal{A}_2$ determines as as not exhibiting $\mathcal{P}$. This disagreement contributes to the vagueness of $\mathcal{P}$. 

As stated earlier, we can think of differences in $\{\mathcal{S}\}_C$ and $\{\mathcal{S}\}_{\bar{C}}$ as differences in $\mathcal{P}$ itself between $\mathcal{A}_1$ and $\mathcal{A}_2$ (This difference can be trivially dissolved in the case above by making $ \{\mathcal{S}\}^{\mathcal{A}_2\mathcal{P}}_{\bar{C}} = \{\mathcal{S}\}^{\mathcal{A}_1\mathcal{P}}_{\bar{C}} = \{ \}$). We can see this clearly with the following example - for property $\mathcal{P}$ we will assume that both observers share the same $\{\mathcal{S}\}^{\mathcal{P}}_C$, and  $\{\mathcal{S}\}^{\mathcal{A}_1\mathcal{P}}_{\bar{C}} = \{ \}$ and $ \{\mathcal{S}\}^{\mathcal{A}_2\mathcal{P}}_{\bar{C}} \neq \{ \}$ for $\mathcal{A}_1$ and $\mathcal{A}_2$ respectively. Both observers use the same function $\phi_{\mathcal{P}}$ which is bounded as: $0 \leq \phi_{\mathcal{P}} \leq \beta_{\phi}$. Both observers will obtain the same $\eta_0^{\mathcal{P}}$ such that any system $\mathcal{S}$ with $\phi_{\mathcal{P}} \geq \eta_0^{\mathcal{P}}$ is considered to be exhibiting $\mathcal{P}$. The interval of $\phi_{\mathcal{P}}$ for borderline cases is $(0,\eta_0^{\mathcal{P}})$ and $(\gamma_0^{\mathcal{A}_2\mathcal{P}},\eta_0^{\mathcal{P}})$ for $\mathcal{A}_1$ and $\mathcal{A}_2$ respectively. Now consider a property $\mathcal{P}^{\prime}$ constructed for $\mathcal{A}_2$ such that $\{\mathcal{S}\}^{\mathcal{P}^{\prime}}_C = \{\mathcal{S}\}^{\mathcal{P}}_C \cup \{\mathcal{S}\}^{\mathcal{A}_2^\mathcal{P}}_{\bar{C}}$ and $\{\mathcal{S}\}^{\mathcal{P}^{\prime}}_{\bar{C}} = \{ \}$. We now construct $\psi_{\mathcal{P}^{\prime}}$ in such a manner that for any $\mathcal{S}_i$ with $\phi_{\mathcal{P}}(\bar{x}^{\mathcal{S}_i} ) \leq \gamma_0^{\mathcal{A}_2\mathcal{P}}$ and $\phi_{\mathcal{P}}(\bar{x}^{\mathcal{S}_i} ) \geq  \eta_0^{\mathcal{P}}$, the corresponding values for $\psi_{\mathcal{P}^{\prime}}(\bar{x}^{\mathcal{S}_i} )$ is appropriately scaled to lie in the interval $[\eta_0^{\mathcal{P}},\beta_{\phi}]$. Also for any  $\mathcal{S}_i$ with $\gamma_0^{\mathcal{A}_2\mathcal{P}}< \phi_{\mathcal{P}}(\bar{x}^{\mathcal{S}_i} ) < \eta_0^{\mathcal{P}}$, the corresponding values of $\psi_{\mathcal{P}^{\prime}}(\bar{x}^{\mathcal{S}_i} )$ is mapped into the interval $(0,\eta_0^{\mathcal{P}})$. Under this construction, $\psi_{\mathcal{P}^{\prime}}$ has the same lower and upper bound of $0$ and $\beta_{\phi}$ as $\phi_{\mathcal{P}}$, and $\eta_0^{\mathcal{P}^{\prime}}=\eta_0^{\mathcal{P}}$. We also get $\gamma_0^{\mathcal{A}_2\mathcal{{P}^{\prime}}}=\gamma_0^{\mathcal{A}_1\mathcal{P}}=0$. The determination intervals for $\mathcal{A}_2$ with respect to $\mathcal{P}^{\prime}$ using $\psi_{\mathcal{P}^{\prime}}$ will be the same as the one $\mathcal{A}_1$ uses for $\mathcal{P}$ using $\phi_{\mathcal{P}}$ for any system $\mathcal{S}$ i.e. $\mathcal{S}$ exhibits $\mathcal{P}^{\prime}$ if $\psi_{\mathcal{P}^{\prime}} \geq \eta_0^{\mathcal{P}}=\eta_0^{\mathcal{P}^{\prime}}$, is borderline if $\psi_{\mathcal{P}^{\prime}}$ is in $(0,\eta_0^{\mathcal{P}})$ and does not exhibit $\mathcal{P}^{\prime}$ only when $\psi_{\mathcal{P}^{\prime}} = \psi_{\mathcal{P}}= 0$. The construction of $\psi_{\mathcal{P}^{\prime}}$ is based only on the clear cases in $\{\mathcal{S}\}^{\mathcal{P}^{\prime}}_C$, and we see that $\mathcal{P}$ and $\mathcal{P}^{\prime}$ do not correspond to the same property as they are systems $\mathcal{S}$ for which $\mathcal{A}_1$ and $\mathcal{A}_2$ would make different determinations with respect to $\mathcal{P}$ and $\mathcal{P}^{\prime}$. This further reinforces that the differences in $\{\mathcal{S}\}_C$ and $\{\mathcal{S}\}_{\bar{C}}$ does result in a different description of $\mathcal{P}$ and $\phi$. In the next section, we will explain how the vagueness discussed under our model lays the path for pan-nifty-ism.

\section{From Vagueness to Pan-nifty-ism}
The title and the content of the paper have been influenced by the quote from Dennett on \textit{pan-nifty-ism} \cite{Dennett} - \textit{``Everything is nifty. Electrons are nifty. Protons are nifty. Molecules are nifty. Pan-nifty-ism.''} While Dennett was making an argument against the metaphysics of panpsychism, we would like to state clearly that the work here cannot and does not take a position on any metaphysical stance. However our metaphysical stances are influenced by epistemic information (which includes that provided by a quantification function $\phi$), and quoting Massimo Pigliucci - \textit{``..one’s metaphysical claims should never be too far from one’s epistemic warrants for those claims''} \cite{Massimo}. Dennett's quote thus inspired this present inquiry into the conditions under which the quantification of $\mathcal{X}$ by $\phi$ produces results that would push one towards a metaphyiscal stance of pan-$\mathcal{X}$-ism, backed by the epistemic warrants obtained from $\phi$. Under the right conditions, when $\mathcal{X}$ is 'nifty', we can go from the quantification of nifty to pan-nifty-ism. 

We can build off the special case discussed at the end of the last section. When two observers $\mathcal{A}_1$ and $\mathcal{A}_2$ share the same $\{\mathcal{S}\}_C$, with $\{\mathcal{S}\}^{\mathcal{A}_1}_{\bar{C}} = \{ \}$ and $ \{\mathcal{S}\}^{\mathcal{A}_2}_{\bar{C}} \neq \{ \}$ and have the same $\phi_{\mathcal{X}}$ for the property $\mathcal{X}$ (not assumed to either sharp or vague apriori), we know that $\mathcal{A}_1$ and $\mathcal{A}_2$ will both agree on which $\mathcal{S}$ exhibit $\mathcal{X}$ according to $\phi_{\mathcal{X}}$. We also have that for $\mathcal{A}_1$, a system does not exhibit $\mathcal{X}$ only when $\phi_{\mathcal{X}}=\alpha^{\mathcal{X}}$ and will be considered borderline for all values of $\phi_{\mathcal{X}}$ in the interval $(\alpha^{\mathcal{X}}, \eta_0^{\mathcal{X}})$. Now let us assume that in addition to being a faithful quantification, $\phi_{\mathcal{X}}$ has been favorably constructed to quantitatively distinguish between systems $\mathcal{S}_i$ and $\mathcal{S}_j$ in $\{\mathcal{S}\}^{\mathcal{X}}_C$ that matches with our expectations of $\mathcal{X}$. For example, we have $\mathcal{S}_i$ smaller than $\mathcal{S}_j$ in scale and the observer's description of $\mathcal{X}$ is such that we get $\phi(x^{\mathcal{S}_i}) << \phi(x^{\mathcal{S}_j})$ as expected. If extremely small systems $\mathcal{S}$ gave low values of $\phi(x^{\mathcal{S}})$ such that $\alpha^{\mathcal{X}} < \phi(x^{\mathcal{S}}) < \eta_0^{\mathcal{X}}$, then $\mathcal{A}_1$ would consider $\mathcal{S}$ as a borderline case of $\mathcal{X}$. Furthermore if the low value of $\phi(x^{\mathcal{S}})$ matches with the expectation that smaller systems $\mathcal{S}$ should have lower $\phi(x^{\mathcal{S}})$ values than larger systems, then $\mathcal{A}_1$ is led to the conclusion that even extremely small systems like $\mathcal{S}$ could exhibit $\mathcal{X}$ to a certain amount given by $\phi(x^{\mathcal{S}})$ (This would seem erroneous to observer $\mathcal{A}_2$ with $ \{\mathcal{S}\}^{\mathcal{A}_2}_{\bar{C}} \neq \{ \}$, who would have determined the same $\mathcal{S}$ as a case of not exhibiting $\mathcal{X}$ if $\phi(x^{\mathcal{S}}) \leq \gamma_0^{\mathcal{X}}$). If there are systems at the smallest spatial and temporal scales that fall into the borderline case interval for $\mathcal{X}$, observer $\mathcal{A}_1$ will conclude that since $\mathcal{X}$ is exhibited at the current smallest scales. And these results imply that $\mathcal{X}$ is a fundamental part of nature exhibited by all systems to some degree - hence laying the \textit{inevitable} path to the metaphysical stance of pan-$\mathcal{X}$-ism. A stronger case for $\mathcal{X}$ at the fundamental scale could be made if we have systems at this scale with $\phi(x^{\mathcal{S}}) > \gamma_0^{\mathcal{X}}$ when $ \{\mathcal{S}\}^{\mathcal{A}_2}_{\bar{C}} \neq \{ \}$ and $\gamma_0^{\mathcal{X}} > \alpha^{\mathcal{X}}$. Note that this would be true for any $\mathcal{X}$, if the quantification of $\mathcal{X}$ maps on to the cases described earlier. The authors thus conjecture that the pan-$\mathcal{X}$-ist implications have already been encoded into the structure of the quantification metric.

If $\mathcal{X}$ is consciousness, then we can study the path from the quantification of consciousness towards panpsychism. The study of consciousness is vital and Integrated Information Theory (IIT) \cite{Tononi} is often identified as the first framework to provide a quantification of consciousness through the use of it's $\phi_{IIT}$ metric. It is extremely popular among panpsychists, since it is the 1st scientific theory of consciousness that seems to imply panpsychism. Though IIT is a phenomenology based approach, the use of aspects of human consciousness in the postulates, axioms and derivation of $\phi_{IIT}$ would serve as the clear case elements of $\{\mathcal{S}\}_C^{IIT}$ and be subject to the model in this paper. In IIT, we have $\phi_{IIT}$ bounded as $0 \leq \phi_{IIT} \leq +\infty $, with a system being considered to not exhibit consciousness only if $\phi_{IIT}=0$ and considered to be conscious to varying degree for any $\phi_{IIT}>0$. As a result, IIT assigns consciousness to a much larger set of systems, many of which we might apriori identify as not being conscious. Now since IIT is built upon the elements of $\{\mathcal{S}\}_C^{IIT}$ and does not seem to apriori identify any system as non-conscious, we have $\{\mathcal{S}\}_{\bar{C}}^{IIT}=\{ \} $. As discussed in the previous section, the difference between whether or not $\{\mathcal{S}\}_{\bar{C}}^{IIT}=\{ \}$ implies a difference in the \textit{`consciousness'} quantified by $\phi_{IIT}$ even when $\{\mathcal{S}\}_C^{IIT}$ remains the same in the two cases. In his blogpost about IIT, Scott Aaronson makes the same point \cite{Aaronson} - \textit{`` On reflection, I firmly believe that a two-state solution is possible, in which we simply adopt different words for the different things that we mean by “consciousness”—like, say, $consciousness_{Real}$ for my kind and $consciousness_{WTF}$ for the IIT kind. OK, OK, just kidding!  How about ``paradigm-case consciousness'' for the one and ``IIT consciousness'' for the other.'} The case of IIT and $\phi_{IIT}$ maps well on to the case discussed above and in the previous section. While $\eta_0^{IIT}$ has not been actually calculated, the existence of $\{\mathcal{S}\}_C^{IIT}$ and $\phi_{IIT}$ ensures that there is a $\eta_0^{IIT}$. Thus all the systems $\mathcal{S}$ with $0 < \phi_{IIT} < \eta_0^{IIT}$ now represent borderline cases of exhibiting consciousness (In reality, since a clear $\eta_0^{IIT}$ has not been calculated yet and $\gamma_0^{IIT}=0$, most people simply refer to the system as exhibiting consciousness to certain extent if $\phi_{IIT} >0$). If this includes systems at very small scales, then we can see how the implication of panpsychism arises. On the other hand, if we apriori identify which systems are not conscious and hence have $\{\mathcal{S}\}_{\bar{C}}^{IIT} \neq \{ \}$, we can calculate $\gamma_0^{IIT} > 0$ using $\phi_{IIT}$ and a non-empty $\{\mathcal{S}\}_{\bar{C}}^{IIT}$, and determines systems $\mathcal{S}$ with $\phi_{IIT} \leq \gamma_0^{IIT}$ as non-conscious systems. With the suitable choice of $\{\mathcal{S}\}_{\bar{C}}^{IIT}$ and $\gamma_0^{IIT}$, we might have $\phi_{IIT} \leq \gamma_0^{IIT}$ for extremely small scale systems that will now be determined to be a non-conscious system (and thus not attribute consciousness to the fundamental scales). However with $\gamma_0^{IIT} > 0$, if we were to find systems at the smallest scales with $\phi_{IIT} > \gamma_0^{IIT}$, that would be much stronger justification for adopting the panpsychist metaphysics. It would thus be erroneous to say that results from IIT imply panpsychism, since we can see that these have been already built into the construction of $\{\mathcal{S}\}_C^{IIT}$ and $\{\mathcal{S}\}_{\bar{C}}^{IIT}$ from which $\phi_{IIT}$ is derived.

\section{Discussion}
In this work we built a simple model of the quantification of sharpness and vagueness, produced definitions of both under the model and analyzed the various cases of interest. We also showed that for any $\mathcal{X}$, if the quantification metric is built under some certain assumptions, then this metric and the underlying framework will provide epistemic backing for the metaphysical stance of pan-$\mathcal{X}$-ism. When we set $\mathcal{X}$=`consciousness,' and use $\phi_{IIT}$ from Integrated Information theory within our model, we can lay the path towards panpsychism.  There is nothing unique about consciousness or $\phi_{IIT}$, and will work just as well for $\mathcal{X}=$`nifty' to pan-nifty-ism. The necessity of whether nifty needs to be quantified is beyond the scope of this work.

\bibliographystyle{unsrt}  

\end{document}